\documentclass[letterpaper, 10 pt, conference]{ieeeconf}  
\usepackage{times}
\usepackage{booktabs}
\usepackage[backend=biber]{biblatex}
\usepackage{multicol}
\usepackage[utf8]{inputenc}
\usepackage{graphicx}
\usepackage{lipsum}
\usepackage{wrapfig}
\usepackage[ruled,vlined,linesnumbered,boxed]{algorithm2e}
\usepackage{soul}
\usepackage{todonotes}
\usepackage{amsmath, amssymb}
\usepackage{cleveref}
\usepackage{bm}
\usepackage{color}
\usepackage{multirow}
\usepackage{comment}
\usepackage{float}
\usepackage{caption}
\usepackage{subfigure}
\usepackage{siunitx}
\definecolor{Burgundy1}{RGB}{128,0,32}
\def\imagetop#1{\vtop{\null\hbox{#1}}}

\let\oldnl\nl
\newcommand{\nonl}{\renewcommand{\nl}{\let\nl\oldnl}}

\IEEEoverridecommandlockouts                              

\title{\LARGE \bf
Amortized Q-learning with Model-based Action Proposals for Autonomous Driving on Highways
}

\author{Branka Mirchevska$^{1,2}$, Maria H{\"u}gle$^{1}$, Gabriel Kalweit$^{1}$, Moritz Werling$^{2}$, Joschka Boedecker$^{1, 3}$
\thanks{$^{1}$Dept. of Computer Science, University of Freiburg, Germany.}%
\thanks{{\tt \{hueglem,kalweitg,jboedeck\}@cs.uni-freiburg.de}}
\thanks{$^{2}$BMW Group, Unterschleissheim, Germany.}%
\thanks{{\tt  \{Branka.Mirchevska,Moritz.Werling\}@bmw.de}}%
\thanks{$^{3}$Cluster of Excellence BrainLinks-BrainTools, Freiburg, Germany.}%
}

\begin{document}

\maketitle
\thispagestyle{empty}
\pagestyle{empty}

\begin{abstract}

Well-established optimization-based methods can guarantee an optimal trajectory for a short optimization horizon, typically no longer than a few seconds.
As a result, choosing the optimal trajectory for this short horizon may still result in a sub-optimal long-term solution.
At the same time, the resulting short-term trajectories allow for effective, comfortable and provable safe maneuvers in a dynamic traffic environment.
In this work, we address the question of how to ensure an optimal long-term driving strategy, while keeping the benefits of classical trajectory planning.
We introduce a Reinforcement Learning based approach that coupled with a trajectory planner, learns an optimal long-term decision-making strategy for driving on highways.
By online generating locally optimal maneuvers as actions, we balance between the infinite low-level continuous action space, and the limited flexibility of a fixed number of predefined standard lane-change actions. 
We evaluated our method on realistic scenarios in the open-source traffic simulator SUMO and were able to achieve better performance than the 4 benchmark approaches we compared against, including a random action selecting agent, greedy agent, high-level, discrete actions agent and an IDM-based SUMO-controlled agent.

\end{abstract}

\section{INTRODUCTION}
Autonomous vehicles have to make safe and reliable decisions in highly dynamic and complex traffic environments. So far, the approaches proposed for tackling this challenge can be divided into three broad categories: rule-based systems, optimization-based (optimal control) systems and data-driven i.e machine learning-based systems.
Rule-based systems \cite{Montemerlo2009JuniorTS}, \cite{Urmson2008AutonomousDI}, \cite{Bacha2008OdinTV}, 
offer the advantage of greater control over their actions.
However, defining a consistent set of rules that works in all possible situations under noisy observations is a difficult and error-prone procedure. 
Optimal control-based systems, like \cite{moritz_thrun}, \cite{Schwarting2017ParallelAI}, \cite{Borrelli2005MPCBasedAT} and \cite{Falcone2007PredictiveAS}, which rely on constrained optimization and Model Predictive Control, encode the vehicle dynamics model directly in the planning module and are able to generate feasible and comfortable trajectories. They also typically offer sound solutions with mathematically backed up safety guarantees. However, due to the short optimization horizon, these approaches are not capable of making farsighted, globally optimal decisions \cite{Schwarting2018PlanningAD}.
Alternatively, purely machine learning-based approaches, while offering better generalization to unseen situations and learning from data, introduce safety concerns and reduce the transparency of the behavior of the system. 
Among machine learning methods, Reinforcement Learning (RL) has become increasingly popular in autonomous driving applications, due to the notable success in many application areas \cite{mnih2013playing,Silver2016MasteringTG,Silver2017MasteringTG}. However, when applying RL to autonomous driving, an important challenge is to determine the level of control the agent should have over the vehicle. Most common are the discrete actions and continuous actions settings.
In the discrete actions case (e.g. \cite{alizadeh2019automated, Hoel, You} and also \cite{Mirchevska2018HighlevelDM,Hgle2019DynamicIF,Hgle2020DynamicIS,Kalweit2020InterpretableMT,kalweit2020deep}), the agent can choose from actions such as keep lane, lane-change to the left or lane-change to the right or fixed accelerating/decelerating steps. 
While the small and fixed action set leads to fast learning progress, the lane-change maneuvers are usually with fixed execution duration, resulting in a sub-optimal, unnatural behavior in tight situations.
On the other end of the spectrum, in the continuous actions setting, the agent learns to influence either the low-level steering wheel and gas/brake pedal control signals directly, or some kind of an abstraction thereof such as curvature or acceleration \cite{wang2019deep,wang2018reinforcement,wang2019quadratic}. 
This endows the agent with maximal freedom of choice, but since the set of actions is infinite and the choices that the agent needs to make are typically very granular, it will need large amounts of good action sequences among the learning samples, slowing down learning considerably.
In addition, the maximization of the action-value function in continuous off-policy RL methods is problematic. To address that, in \cite{enormousActions}, the maximization step is simplified by a learned proposal distribution which is used to sample the continuous action space. 
The approximate maximum is then taken as the maximizing action from the proposed sample-set.

\begin{figure}[t]
	\centering
    \includegraphics[scale=0.26]{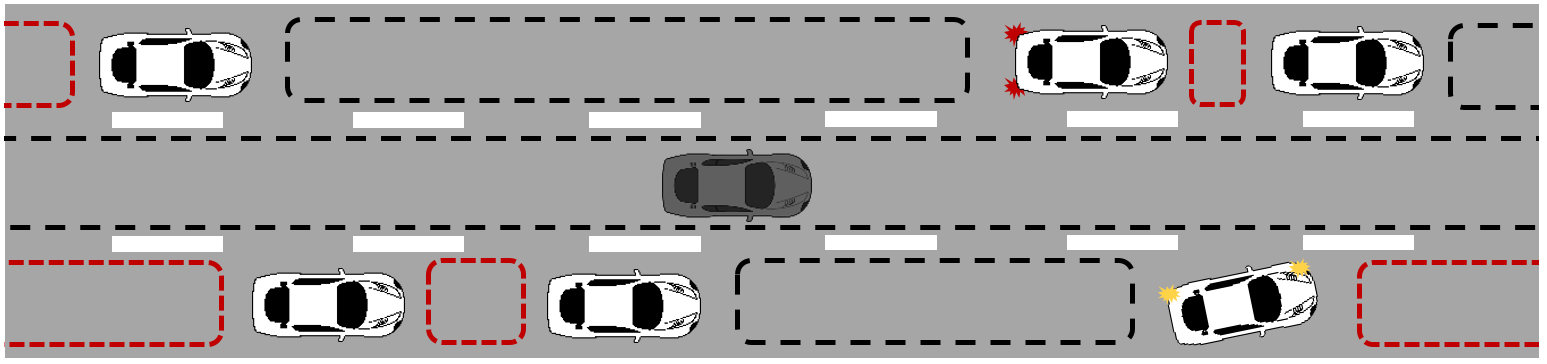}
	\caption{RL environment: RL agent (black), surrounding vehicles (white), reachable gaps (black-dashed), unreachable gaps (red-dashed)}
	\label{environment}
\end{figure}

Inspired by this approach, we use model-based action proposals for balancing out the trade-off between continuous and discrete actions discussed above (see \Cref{environment} for an overview). Since the actions that we are proposing, are finite but described by continuous values, they can be seen as a compromise between the discrete and the continuous representations. 
This way, we combine the advantages of both sides of the spectrum, while ensuring safety. 
For this purpose, we introduce the notion of gaps, as model-based action proposals for the agent.

A gap is a space between two surrounding vehicles on the same lane. We define the actions as the set of all reachable gaps in the vicinity of the agent. 
An embedded trajectory planner plans locally optimal trajectories which satisfy all physical and safety constraints, based on which the set of reachable gaps is created.
After the RL agent chooses a gap from the reachable set of gaps, the best low-level maneuver towards that gap is executed.
Since, the number of gaps the agent can choose from is finite, but the features describing each gap are continuous, we allow for a sufficient level of control over the vehicle, while ensuring safety and keeping the action space and the learning times manageable. 

We introduce \textit{Amortized Q-learning with Model-based Action Proposals} (AQL-MAP), an \textit{Amortized Q-learning} approach that replaces the learned proposal distribution by model-based action estimation. 
The term "amortized" in this case stems from amortizing the cost of maximization over the whole continuous action space by proposing a finite set of reachable actions to maximize over. Even though the set of proposed actions is finite, they cover all safe and feasible maneuvers the agent can execute in a given state.
We implement the approach in the Semi-MDP Options-framework \cite{Sutton1999BetweenMA}, \cite{shalevshwartz2016safe} in order to speed up learning and stabilize the performance.
Moreover, since for autonomous driving in real traffic, on-policy, online learning is out of the question, the training data usually needs to be pre-collected from fleets of human-operated vehicles. 
With that in mind, we train our approach in offline and off-policy fashion, on fixed batch of training data.
Our main contributions are the following:
\begin{enumerate}
	\item Introducing gaps as an action abstraction, described by continuous features which incorporate all possible maneuvers that the vehicle can safely perform.
	\item Coupling RL with an optimization based trajectory planner for estimating the current set of accessible gaps and generating feasible and safe trajectories towards them.
	\item Analysis of the achieved performance by evaluating the novel agent against 4 benchmark agents in realistic highway scenarios, defined in the Sumo simulation environment \cite{sumo}. 
	The benchmark agents include: random action selecting agent, greedy agent, IDM-based Sumo-controlled agent \cite{Treiber_2000} and a high-level, discrete action selecting agent.
\end{enumerate}

\section{REINFORCEMENT LEARNING BACKGROUND}\label{rl}
Reinforcement learning is concerned with the problem of learning from interaction with the environment. 
Formally we define an RL problem as a Markov Decision Process, consisting of: set of states $S = \{s_1, s_2, ..., s_n\}$, set of actions $A = \{a_1, a_2, ..., a_m\}$, transition function $T$: $T(s, a, s') = P(s(t+1) = s'| s(t) = s, a(t) = a)$ and a reward function  $r(s, a)$.
In a state $s_t$, at a time-step $t$, following a policy $\pi$, an RL agent executes an action $a_t$ in the environment. 
At a time-step $t+1$, it then ends up in some state $s_{t+1}$ and receives a reward $r_{t+1}$.
We define $R(s_t) = \sum_{t'>=t} \gamma^{t'-t}r_{t'}$ as the expected discounted long-term return, where $\gamma \in [0, 1]$ is the discount factor. 
The objective of the agent is to optimize its policy $\pi$ in order to maximize $R$.
One way of finding the optimal policy is by Q-learning \cite{Watkins92q-learning}, where the agent aims to learn the optimal action-value function $Q^*(s, a)$, defined as: $Q^*(s, a) = \max_{\pi}\mathbb{E}[R_t|s_t = s, a_t = a, \pi]$.\\

Since we implement our approach in the Options framework, here we briefly introduce it.
Options are a generalization of primitive actions to include temporary extended courses of actions.
An option is a triple $o = (I, \pi, \beta)$, where $ I \subset S$ is the set of states that an option can start in (initiation set), $\pi: S \times A \rightarrow [0, 1]$ is the policy and $\beta: S^+ \rightarrow [0, 1]$ is the termination condition ($S^+$ is the set of all states including the terminal state if there is one).
An example of an option would be the task of changing a lane, which consists of multiple primitive actions like: activating the turn signal, turning the steering wheel, stepping on the gas pedal etc.
A set of options defined over an MDP constitutes a semi-Markov decision process (SMDP).
Options can take variable number of steps.
Given a fixed set of options $O_s$ at state $s$, at a time-step $t$ the agent chooses an option $o_t$.
When the option terminates at time-step $t+k$, the agent is to select a new one (unlike the standard MDP setting when the agent always chooses a new primitive action at $t+1$).
Using options can potentially speed up and stabilize learning on large problems.
For more details refer to \cite{Sutton1999BetweenMA} and \cite{Sutton1998}.

\section{GENERAL APPROACH}\label{method}
The final approach consists of two main components: the trajectory planning and the learning algorithm. 
For generating the set of action proposals for each time-step, we adapted the sampling-based optimal control approach for trajectory planning and generation from \cite{moritz_thrun}.
For approximating the optimal $Q$-function, we use DQN-based function approximator $Q$, with parameters $\theta$, in an offline, batch learning way.
Namely, each training iteration, we sub-sample a mini-batch $M$ of transition samples $(s, a, r, s')$, from a pre-collected training data buffer $R$.
Given $M = (s_i, a_i, s_{i+1}, r_i)$, we optimize for a the loss function: $L(\theta) = \frac{1}{b} \sum_i (Q(s_i, a_i | \theta) - y_i)^2$, where $y_i = r_i + \gamma \max_a Q'(s_{i+1}, a_i|\theta)$ are the targets.
$Q'$ is the target network with parameters $\theta'$ which is updated every iteration by a soft update policy with parameter $\tau \in [0, 1]$.
In order to reduce the overestimation of the $Q$-values we use two target networks, as suggested in \cite{vanhasselt2015deep}. Furthermore, we build on the DeepSet-Q approach from \cite{Hgle2019DynamicIF}, which allows us to support variable number of inputs for our RL architecture. 

\begin{figure}[t]
	\centering
	\includegraphics[scale=0.3]{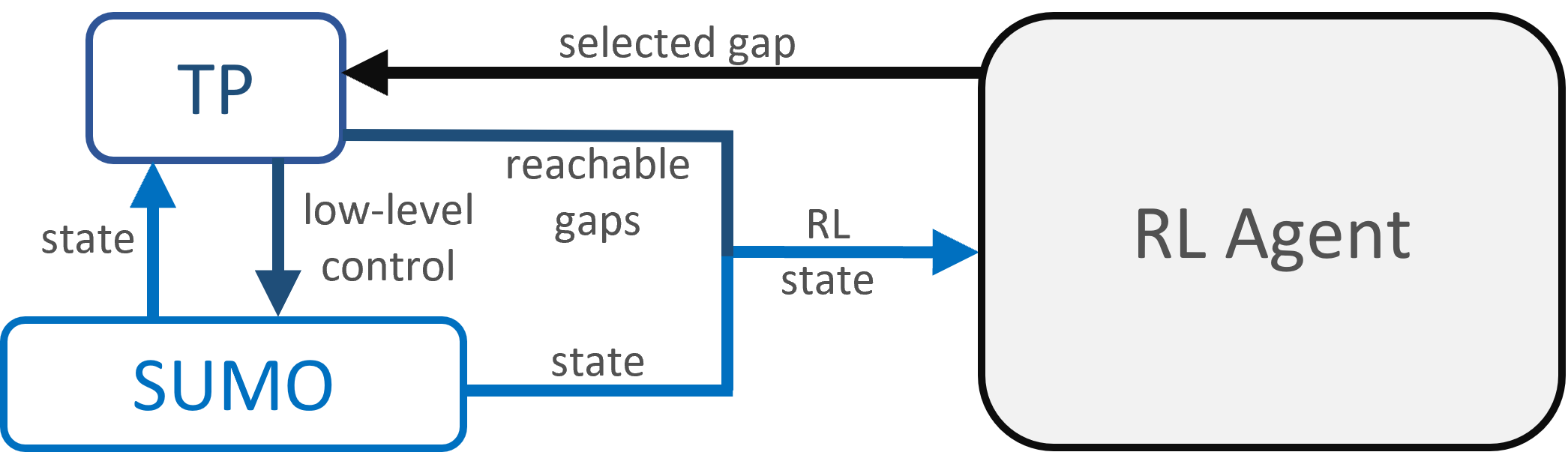}
	\caption{Diagram of the information flow between Sumo, TP and the RL agent}
	\label{rl_env}
\vspace*{-0.4cm}
\end{figure}

\subsection{AQL-MAP Framework}\label{framework}
In \Cref{rl_env} the complete framework of interaction between the simulation environment (Sumo), the trajectory planner (TP) and the RL agent is shown. 
In each time-step, first Sumo sends the current environment state to TP. 
Based on it, TP plans locally optimal trajectories and creates the set of reachable gaps. 
The current environment state and the set of reachable gaps construct the RL state which is sent to the RL agent. 
After the RL agent has chosen a gap to fit into,  
it sends its decision to TP. 
Then, TP sends to Sumo, the low-level control signals necessary for reaching the selected gap. 
Sumo executes the controls and sends the next environment state to TP to begin a new cycle of the same process. 

Using this framework, we collect the training data transitions.
In RL state $s_t$, given the initial state of the RL agent: $s^{rl}_t = \{\text{pos}_t, \text{speed}_t, \text{accel}_t\}$, in time-step $t = 0$
considering the surrounding vehicles in the sensor range of the RL agent, the set of all surrounding gaps $G_t = \{g_{t,0}; g_{t,1}; ...; g_{t,n}\}$ is computed.
Then, the TP generates all possible trajectories starting from state $s^{rl}_t$: $T_t = \{tr_{t,0}; tr_{t,1}; ...; tr_{t,m} \}$.
By examining which gaps are reachable by the generated trajectories, we compute the set of reachable gaps: $RG_t = \{rg_{t,0}; rg_{t,1}; ...; rg_{t,k}\}$.
This set is proposed to the RL agent to choose from. 
The RL agent then randomly chooses a gap $g_t \in RG$, upon which the TP returns the best trajectory leading to $g_t$.
The simulator starts executing the trajectory, the RL agent arrives in state $s_{t+1}$ and receives a reward from the environment $r_t$.
Finally, the transition $(s_t, g_t, s_{t+1}, r_t)$ is stored to the fixed batch $\mathcal{R}$.

\subsection{Trajectory Planning}

We plan locally optimal trajectories to each gap if reachable within a \SI{6}{\second} horizon, that satisfy physical and safety constraints.
This is done based on the collision checking module of the TP, as well as some predefined vehicle dynamics constraints like maximum/minimum acceleration and maximal allowed speed.
First, we compute all gaps in the surroundings of the agent.
Then, we check which ones are safely reachable by at least one trajectory.
In this manner, we define the set of reachable gaps, that are proposed to the RL-agent as actions (\Cref{environment}). 
Once the agent selects a target gap, out of all trajectories that safely reach the gap, the TP outputs the best one.
The best trajectory is selected based on a set of costs and weights taking into consideration the speed, the comfort, as well as how well the end points of the trajectory fit inside the gap.
More precisely, we are minimizing the cumulative longitudinal and lateral jerk of the trajectory for comfort. 
In lateral direction, we penalize the lateral distance to the target lane center.
In longitudinal direction, we predict constant velocity trajectories for all relevant\footnote{For e.g. if we are examining a keep-lane trajectory, only the vehicles on that lane are considered relevant.} surrounding vehicles, so that we can maintain safe distance between them and the RL agent. 
This is done by calculating the time-to-collision and time-headway values for the planned trajectory, relative to the relevant surrounding vehicles.
For a host vehicle $v_h$, a reference vehicle $v_r$, we define time-to-collision and time-headway as:
$\text{ttc} = |v_{h, \text{pos}} - v_{r, \text{pos}}| / (v_{f, \text{speed}} - v_{l, \text{speed}})$ and $\text{thw} = |v_{h, \text{pos}} - v_{r, \text{pos}}| / v_{f, \text{speed}}$,
where $v_{h, \text{pos}}$ and $v_{r, \text{pos}}$ are the longitudinal positions of the host and the reference vehicle and, $v_{\text{l, speed}}$ and $v_{\text{f, speed}}$ refer to whichever vehicle is leading/following between the host and the reference vehicle in the time of the calculation.

For more details on the trajectory generation, refer to \cite{moritz_thrun}.

\subsection{Model-based Action Proposals Algorithm}\label{mbap}

\begin{figure}[b]
	\centering
	\vspace*{-0.3cm}\includegraphics[scale=0.25]{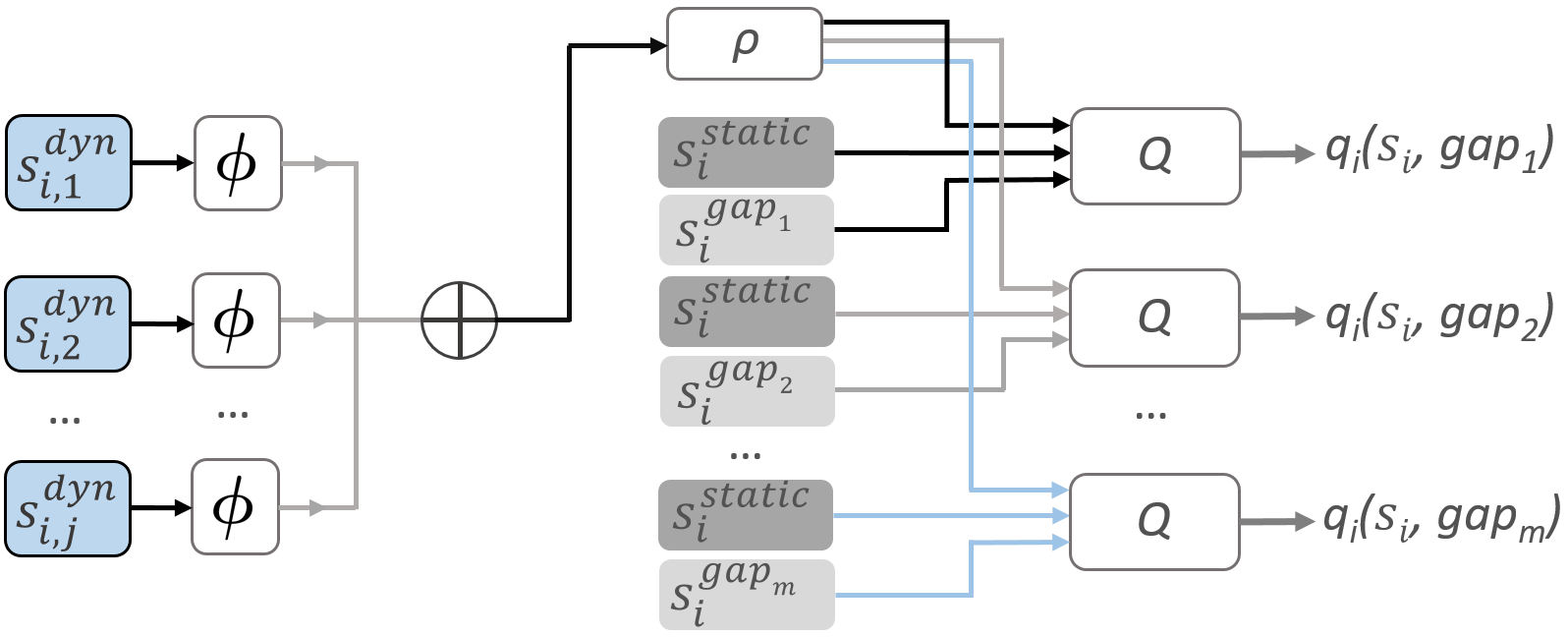}
	\caption{Amortized Q-learning with Model-based Action Proposals network architecture}
	\label{deep_sets_action}
\end{figure}

We use DQN for the state-action value function approximation. 
For that purpose, we initialize a network $Q_{\mathcal{AQL}}$, with parameters $\theta$ and a target network $Q'_{\mathcal{AQL}}$, with parameters $\theta'$.
$Q_{\mathcal{AQL}}$ consists of the modules $\phi$, $\rho$ and $Q$, shown in \Cref{deep_sets_action}.
The fully connected networks $\phi$ transform each of the dynamic input instances $\{s^{\text{dyn}}_{i, 1}, s^{\text{dyn}}_{i, 2}, ..., s^{\text{dyn}}_{i, j}\}$, describing the surrounding vehicles of the RL agent, into the representation $\phi(s^{\text{dyn}}_i)$ as in \cite{Hgle2019DynamicIF}. 
Then a pooling operation (sum) is applied over the output, ultimately making the $Q$ function permutation invariant wrt. its input.
This is what enables handling dynamic number of input objects \cite{NIPS2017_6931}.
The output of the pooling operation is processed by the fully connected network $\rho$, providing the final reconstruction of the dynamic inputs.
Besides the dynamic input instances, we also have a static part of the state $s^{static}_i$ consisting of features describing the RL agent.
In order to inform the RL agent about the available gaps, we include the features describing the reachable gaps.
Finally we combine the final representation of the dynamic inputs $\rho(\sum_{i}\phi(s^{\text{dyn}}_i))$, with as many pairs $(s^{\text{static}}_i, s^{\text{gap}}_i)$ as we have available gaps in time-step $i$.
Then we propagate each triple $(\rho(\sum_{i}\phi(s^{\text{dyn}}_i)), s^{\text{static}}_i, s^{\text{gap}}_i)$ through the $Q$ module separately, to finally get the state-action value estimation $q$ for each proposed gap.
In the end, we acquire the final policy by maximizing over the $q$-value estimations of all proposed gaps, and choosing the one with the maximal $q$-value.
For details refer to \Cref{alg:aql-map}. Building upon \cite{Hgle2019DynamicIF}, the parts marked with red in lines 11 and 12 in \Cref{alg:aql-map}, are the ones that we introduce for the description of the gap and for enabling variable number of actions in each time-step.
Once we have a trained agent, we can apply it on a new driving scenario. The gap selection steps of a trained agent are summarized in \Cref{alg:execution}.

\begin{algorithm}[thpb]
	\SetAlgoLined
	\DontPrintSemicolon
	initialize $Q_{\mathcal{AQL}} = (Q, \bm{\phi}, \rho)$, $Q'_{\mathcal{AQL}} =(Q', \bm{\phi}', \rho')$\\
	set replay buffer $\mathcal{R}$\\
	
	\For{\text{training iteration} $ti=1,2,\dots$}{
		\mbox{get mini-batch $M = (s_i, g_i, s_{i+1}, r_{i+1})$ from $\mathcal{R}$},\\
		where $s_i = (S_i^{\text{dyn}}, s_i^{\text{static}})$, $s_{i+1} = (S_{i+1}^{\text{dyn}}, s_{i+1}^{\text{static}})$ \\
		\ForEach{\text{transition}}{ 
			\ForEach{\text{object} $s_{i+1}^{j}$ in $S_{i+1}^{\text{dyn}}$}{
				$(\phi'_{i+1})^j=\phi'\left(s_{i+1}^j\right)$\\
			}
			$\rho'_{i+1}=\rho'\left(\sum\limits_j(\phi'_{i+1})^j\right)$\\

			 $y_i= $\nolinebreak $r_{i+1}+\gamma\max_a Q'(\rho'_{i+1}, (s_{i+1}^{\text{static}},$ \nolinebreak \vspace{0.3cm}{\textcolor{Burgundy1}{$s_{i+1}^{\text{gap}}$}$), g)$}, \\
			 \textcolor{Burgundy1}{where: $g = s_{i+1}^{\text{gap}} \in \{s_{i+1}^{\text{gap}_1}, s_{i+1}^{\text{gap}_2} \text{, ... , } s_{i+1}^{\text{gap}_m} \}$}\\
		}
		perform a gradient step on loss: $$\frac{1}{b}\sum\limits_i\left(Q_{\mathcal{AQL}}(s_i, a_i)-y_i\right)^2$$\\
		update target network by:
		$$\theta^{Q'}\leftarrow\tau\theta^{Q} + (1-\tau)\theta^{Q'}$$
	}
	\caption{Fixed Batch AQL-MAP}
	\label{alg:aql-map}
\end{algorithm}

\begin{algorithm}[t]
	\SetAlgoLined
	\DontPrintSemicolon
	\nonl \noindent \hspace{-0.4cm} \textbf{Input}: RL agent's state $s^{rl}_t = \{\text{pos}_t, \text{speed}_t, \text{accel}_t\}$, \\
	\nonl \noindent \hspace{0.63cm} RL state $s_t = (s_t^{\text{dyn}}, s_t^{\text{static}})$ in time-step $t = 0$  \\
	\nonl \noindent \hspace{0.63cm} Learned model $Q_{\mathcal{AQL}} = (Q, \bm{\phi}, \rho)$  \\
	
	\While{\text{episode not finished}}{
    get the reachable gap proposals from the TP: \\
    $s_t^{\text{gap}} = \{s_{t}^{\text{gap}_1}, s_{t}^{\text{gap}_2} \text{, ... , } s_{t}^{\text{gap}_m}\}$ \\
    compute the dynamic state representation as in:
	   $\text{state}\_\text{rep}_t^{\text{dyn}} =$ $\rho\left(\sum(\phi_{t}(s_t^{\text{dyn}})\right)$\\
	$\text{Q\_values}_t = \{\}$ \\
	\ForEach{ gap $g_t^m = s_{t}^{\text{gap}_m}$ in $s_t^{\text{gap}}$ }{
	   $\text{Q\_values}_t.\text{insert}(Q_{\mathcal{AQL}}(\text{state}\_\text{rep}_t^{\text{dyn}}, s^{\text{static}}_t, x^{gap}_t))$ \\
	}
	choose gap $g_t = \text{argmax}(\text{Q\_values}_t)$ \\
	}
	\caption{AQL-MAP Learned Agent\newline Gap Selection}
	\label{alg:execution}
\end{algorithm}

\section{TRAINING}\label{application}

We apply the AQL-MAP approach to teach an agent to drive on a simulated highway, in a smooth and safe way, while respecting the desired speed requirement. 
In this section, we discuss the training pipeline and its components.
\subsection{Options Framework}
At a time-step $t$ the RL agent, chooses a gap $g$ (an option) from the set of reachable gaps. 
Unlike the classical RL framework, where the RL agent chooses a new action in fixed time-intervals, here, the duration of the intervals is arbitrary.
Every second, if the gap is not reached and it is still available, the RL agent is not required to choose again.
If the initially chosen option is not available anymore, we can interrupt it by choosing a new, reachable one.
\subsection{State space} \label{state}
The RL state consists of features describing the RL agent, its surrounding vehicles and gaps. For the RL agent we use:
\begin{itemize}
	\item absolute velocity, $v_{\text{RL}}\in\mathbb{R}_{\geq0}$
	\item left lane valid flag, indicating whether there is a lane left of the RL agent, $\text{ll}_{\text{valid}} \in \{0, 1\}$,
	\item analogously for the right lane: $\text{rl}_{\text{valid}} \in \{0, 1\}$
\end{itemize}
\noindent For describing the vehicles $j$ surrounding the RL agent, we consider all vehicles in sensor range $\text{sr}$ of \SI{80}{\meter} ahead and behind. 
We describe them with the following features:\begin{itemize}
	\item relative distance $d_{\text{rel}, j}$, defined as $(\text{pl}_{j} - \text{pl}_{\text{RL}}) / \text{sr}$, where $\text{pl}_{\text{RL}}$ and $\text{pl}_{j}$ are the longitudinal positions of the RL agent and the considered vehicle $j$ respectively.
	\item relative velocity $v_{\text{rel}, j}$, defined as $(v_{j} - v_{\text{RL}}) / v_{\text{des}, \text{RL}}$, where $v_{j}$ is the absolute velocity of the vehicle $j$, and $v_{\text{des, RL}}$ is the user-defined desired velocity the RL agent is aiming to achieve.
    \item relative lane $\text{lane}_{\text{rel}, j}$, defined as $\text{lane}_{\text{ind}, j} - lane_{ind, RL}$, where $\text{lane}_{\text{ind}} \in \{0, 1, 2\}$, represents the right, middle and left lane indices.
\end{itemize}
\noindent Additionally, we include the features for every reachable gap in the surroundings, described below.

\subsection{Action space} \label{action_space}
The action set consists of all reachable gaps $RG$, described in \Cref{framework}. For description of the gaps we use:

\begin{itemize}
    \item relative distance $d_{\text{rel}, g}$, relative velocity $v_{\text{rel}, g}$ and relative lane $\text{lane}_{\text{rel}, g}$, defined the same way as in \Cref{state} for the surrounding vehicles, here in terms of gaps.
	\item gap length, $\text{len}_{g}$, defined as $\text{pl}_{\text{lv}} - \text{pl}_{\text{fv}}$, where $\text{pl}_{\text{lv}}$ and $\text{pl}_{\text{fv}}$ are the longitudinal positions of the leading and the following vehicles forming the gap $g$.
	\item gap association flag, $\text{af}_g \in \{0, 1\}$, which is 0 if the currently chosen gap is the same as the one chosen at the previous time-step, and 1 otherwise. 
\end{itemize}
\subsection{Reward function} \label{reward}
For a desired velocity $v_{\text{des, RL}}$, given $\delta_\text{vel} = |v^{\text{RL}}_s - v_{\text{des, RL}}|$, the reward function $r: S \times A \rightarrow \mathbb{R}$, is defined as :
\begin{equation}
{r(s, a)=\begin{cases}
1 - \delta_\text{vel} + p_{\text{ac}}, & \text{if}\hspace{0.1cm} v^{\text{RL}}_s < v_{\text{des, RL}}.\\
1 + p_{\text{ac}}, & \text{otherwise}.
\end{cases}}
\end{equation} 
where $p_{\text{ac}}$ is a slight penalty for choosing an action that differs from the one chosen in the previous time-step, which encourages smoother driving. 
We use  $p_{\text{ac}} = -0.01$, chosen empirically by conducting a wide range of experiments.

\subsection{Training data details} \label{training_data}

We collected two training data sets, one for the high-level agent, where the RL agent chooses an action every second, and one for the options-selecting agent, which chooses a new option in intervals with varying size (depending on when a selected option terminates).
However, apart from the action/option choosing frequency, we made sure that they are collected in the same way, so everything we describe below holds for both of them. 
In the Sumo simulation environment, we collected around $5 * 10^5$ transition samples $(s, a, s', r)$, that correspond to approximately 150 hours of driving.
The data is collected on a 3-lane straight highway. 
Besides the RL agent, there are between 0 and 70 surrounding vehicles, positioned randomly on the highway, with varying desired speeds and driver behaviors.
\noindent All surrounding vehicles are allowed to change lanes and accelerate/decelerate as their underlying controller suggests, we have control only over the RL agent. 
For the options-selecting agent, we used a random data collection policy.
For the high-level agent, we collected data in a pseudo-random way, where sometimes the agent would keep the previously chosen action with higher probability.
This is in order to make sure that we incorporate samples in the data that describe completely executed lane-changes.

\section{EXPERIMENTS}\label{results}
\subsection{Benchmark agents}

As already mentioned, we compared the performance of the option-selecting RL agent to the one of 4 other benchmark agents: random agent, greedy agent, high-level RL agent and IDM-based Sumo agent.
The random agent is the simplest baseline providing us with a benchmark performance of a non-intelligent agent. 
The greedy agent, out of all reachable gaps, selects the gap with the fastest trajectory leading towards it.
The IDM-based agent is a Sumo-controlled agent.
We set its parameters as relaxed as possible in order to make sure that no internal constraints prevent it from achieving the desired speed (such as no overtaking from the right, being too cooperative etc.). 
Finally, the high-level RL agent chooses from discrete actions: keep lane, left lane-change and right lane-change. Since for this agent there is no need to feed the actions into the network as a part of the RL state, (because they are discrete and fixed,) we used the classic DeepSet DQN architecture. However, we made sure that everything else is comparable to the way the option-selecting agent was trained. We used the same reward function and the same RL state (excluding the options-related features).
During evaluation, all agents were rewarded based on the same criteria. 
The underlying trajectory planning and selecting mechanism is identical for all agents. 
They only differ in the action/option selection strategy.

\begin{figure}[t]
    \centering
    \includegraphics[scale=0.27]{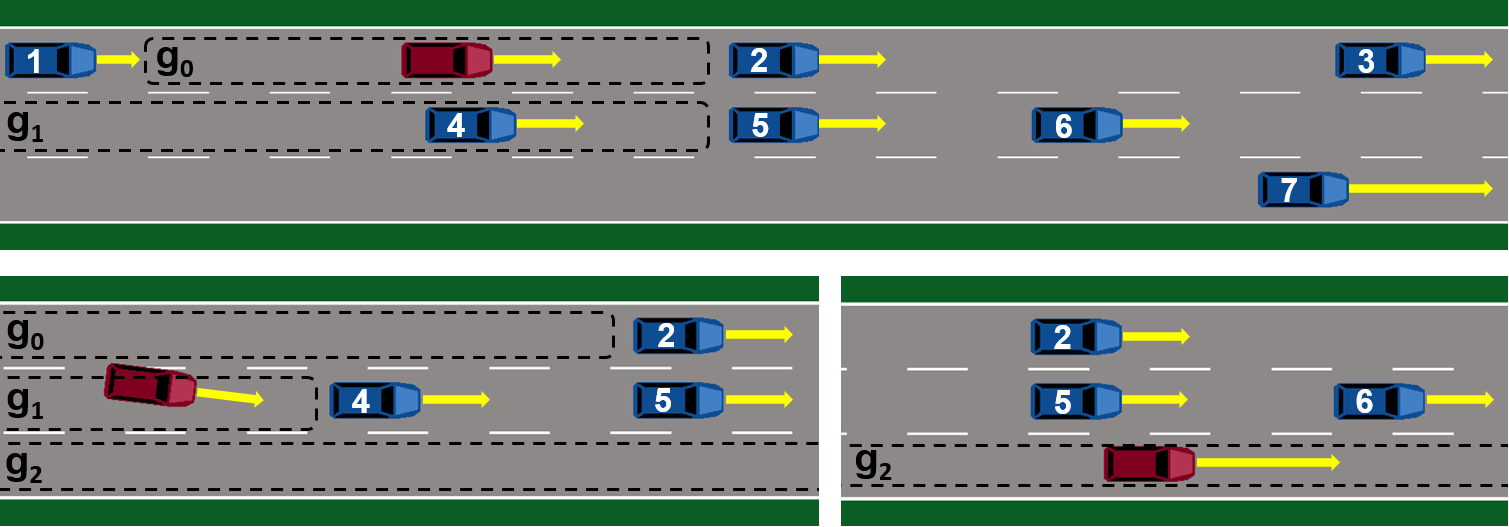}
    \caption{The RL agent (red), surrounded by slow vehicles learned that it is beneficial to move over to the middle lane in order to reach the right-most lane where it will make progress faster.}
    \label{7_cars}
\end{figure}
\begin{figure}[t]
    \centering
    \includegraphics[scale=0.27]{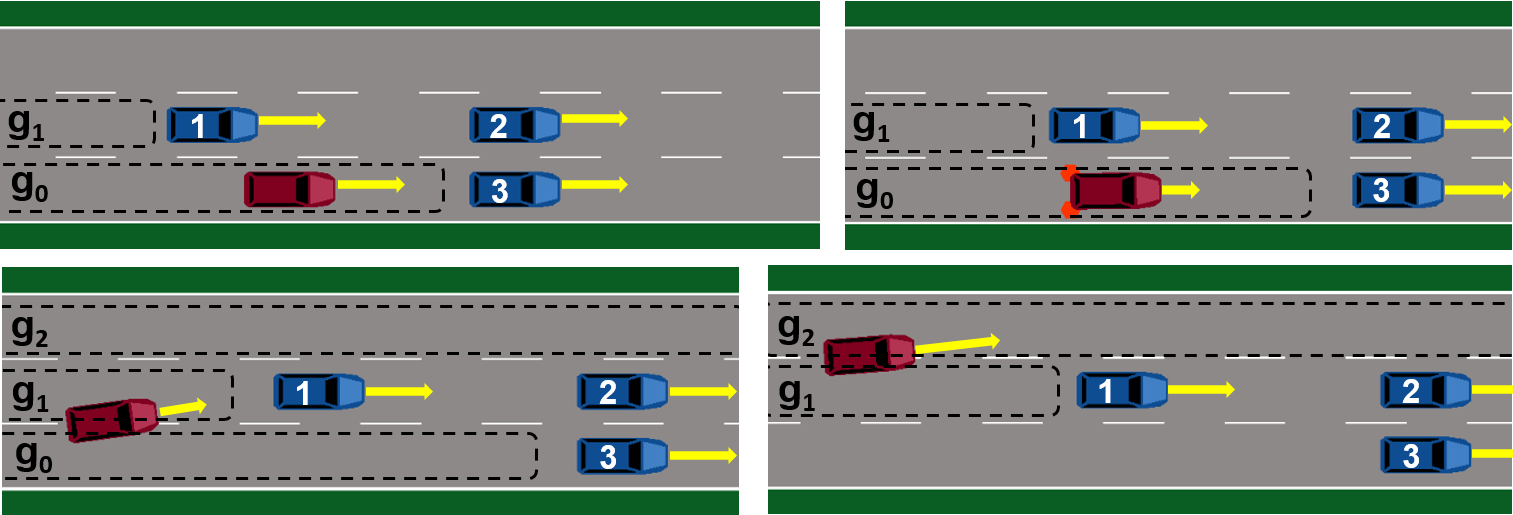}
    \caption{More challenging dense traffic highway situation that can occur for e.g. by entering a highway. Only the options-agent has learned that even though when choosing option $g_1$, prolonged braking maneuver must be applied, it is still the best long-term strategic choice.}
    \label{3_cars}
    \vspace*{-0.6cm}
\end{figure}

\subsection{Evaluation setup}
We evaluated all agents on the same set of realistic scenarios, generated randomly, as described in \ref{training_data}. 
In order to objectively assess the driving of the agents, we generated scenarios with varying number of surrounding vehicles $n = \{10, 20, ..., 80\}$, simulating fluctuating traffic density, from light to heavy. 
For each number of surrounding vehicles we have 10 different evaluation scenarios (80 in total).
Additionally, we evaluate the performance of the agents on critical, more challenging highway scenarios, shown in \Cref{7_cars} and \Cref{3_cars}, where planning ahead is crucial.
We compared the average achieved speeds of all agents on the 80 scenarios. 
The average is calculated over the 10 scenarios with same number of cars in it. 
In order to acquire a better overview of the performance, for the random agent and the RL agents, we calculated an average over 10 different runs.
For the RL agents, we calculated the average over 10 models trained with the same randomly initialized hyper-parameters (\Cref{table_final_params}).
This allows for a better inspection of the robustness of the approach.
We performed over 500 random search runs (configuration space shown in \Cref{table_final_params}) using the Hyperband framework for distributed hyper-parameters optimization \cite{DBLP:journals/corr/LiJDRT16}. 
\begin{table}[ht]
\caption{Training hyper-parameters and configuration spaces}
\begin{center}
\begin{tabular}{ccc}
    \hline
    Parameter & Final value & Search configuration space \\
    \hline
    training steps & $75K$ & \\
    batch size & $64$ & \\
    discount factor & $0.99$ & \\
    update type & soft & \\
    learning rate & $10^{\text{-}4}$ & $[10^{\text{-}1}, 10^{\text{-}2},..., 10^{\text{-}6}]$\\
    $\tau$ & $10^{\text{-}4}$ & $[10^{\text{-}1}, 10^{\text{-}2},..., 10^{\text{-}6}]$\\
    $\phi$ hidden dim. & 20 & $[10, 20, ..., 100]$\\
    $\phi$ output dim. & 80 & $[50, 60, ..., 130]$\\
    $\rho$ hidden dim. & 80 & $[50, 60, ..., 130]$\\
    $\rho$ output dim. & 20 & $[10, 20, ..., 100]$\\
    $\text{FC}_1$/$\text{FC}_2$ dim. & 100 & $[50, 80, 100, 200, 300]$\\
    pooling fn. & $\sum$\\
    \hline
\end{tabular}
\end{center}
\label{table_final_params}
\end{table}

\begin{figure}[thpb]
	\centering
	\hspace*{-0.4cm}\includegraphics[scale=0.5]{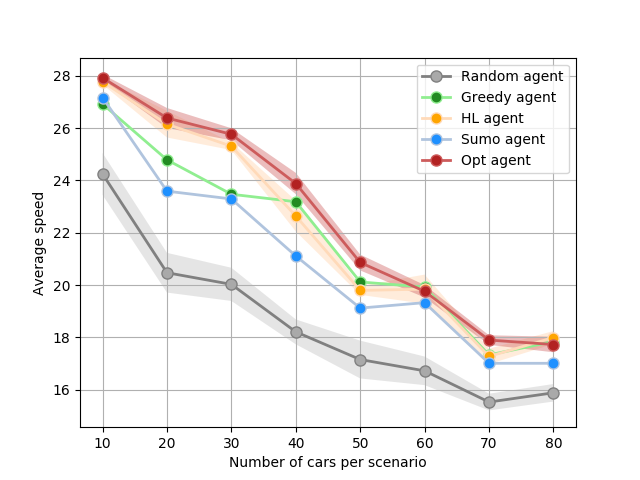}
	\caption{Performance of all agents in the randomly generated scenarios. The shaded areas represent the standard deviation over the 10 runs for the random, the high-level and the options agents.}
	\label{agent_comparisson_speeds}
	\vspace*{-0.6cm}
\end{figure} 

\section{Results}
\Cref{agent_comparisson_speeds} shows the final average speed achieved by each agent in the randomly generated scenarios. 
As expected the random agent (grey) shows the worse performance. 
The greedy agent performs better than the random but worse than the high-level agent and the options-agent.
The Sumo-agent, due to its simplistic strategy falls slightly behind the greedy agent.
The high-level agent and the options-agent have similar performance in the generic scenarios.
This is because in simple, usual scenarios where the others behave predictably, the benefits of long-term planning are less visible.
Additionally, the high-level agent has learned to exploit the longitudinal trajectories of the trajectory planner which is sufficient for navigating the common, randomly generated scenarios.
However, in the section below we show that the options agent is superior as a whole, since when it comes to delicate situations, it has learned to incorporate long-term planning, unlike any of the comparison agents.

\subsection{Performance in the critical scenarios}
\begin{figure}[t]
\centering
\hspace*{-0.4cm}\begin{tabular}{cc}
\centering
 a) Average speed & b) Average duration \\
  \imagetop{\includegraphics[scale=0.25]{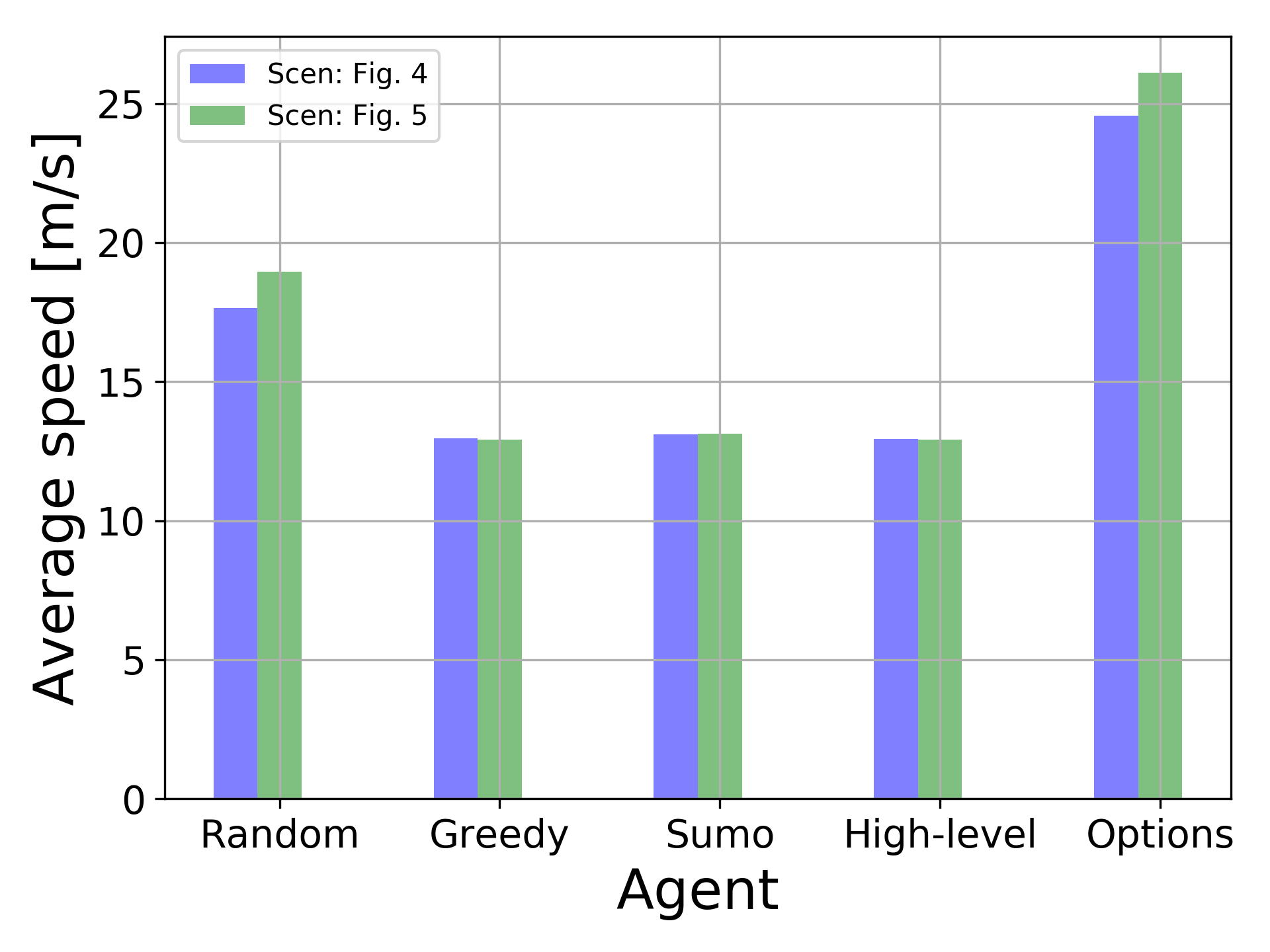}} &
  \imagetop{\includegraphics[scale=0.25]{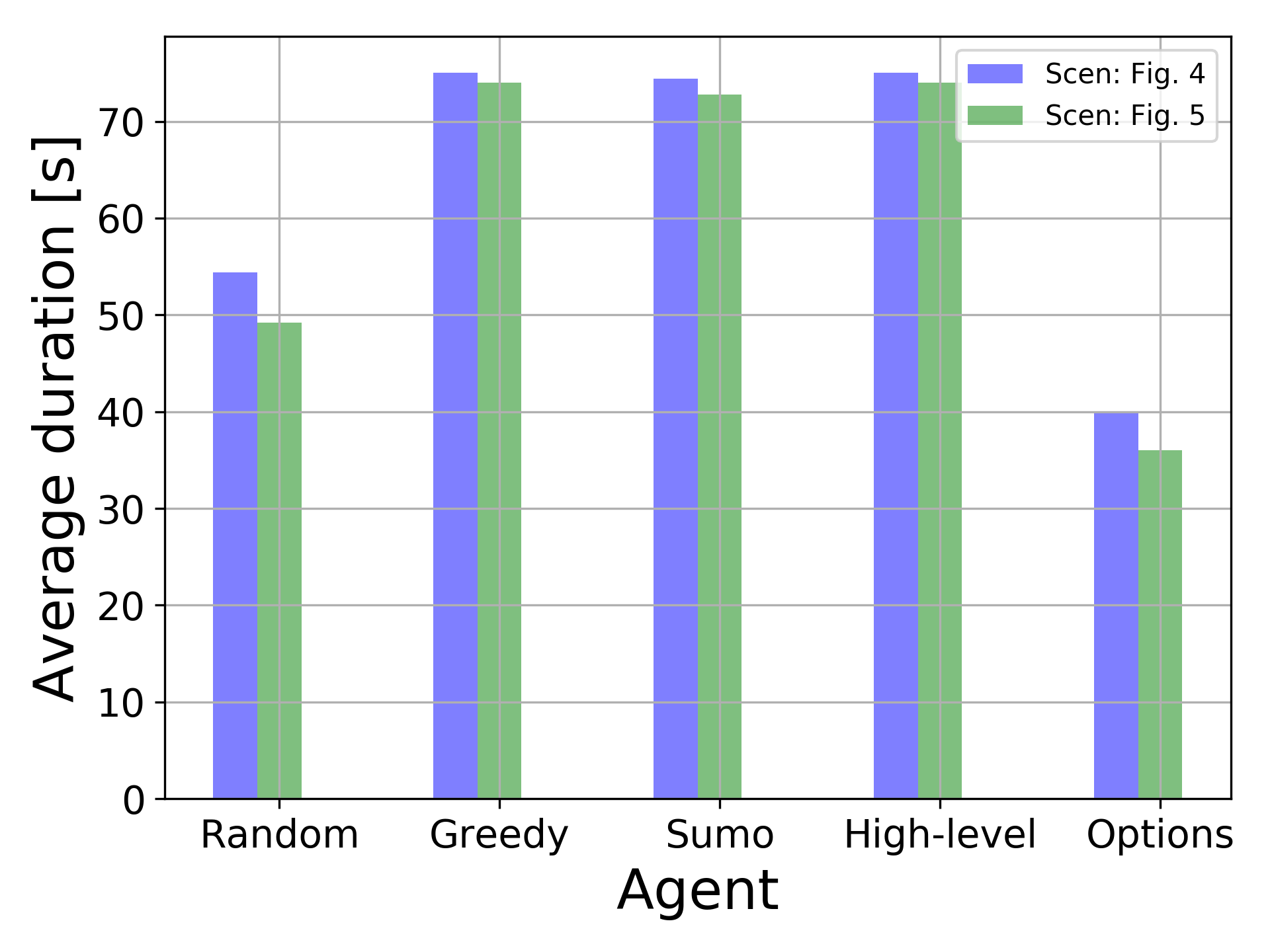}} 
\end{tabular}
\caption{Performance of all agents in the challenging scenarios from \Cref{7_cars} and \Cref{3_cars} in terms of speed and duration.}
\label{critical_scen}
\vspace*{-0.6cm}
\end{figure}

For further investigation of the performance, we looked into specific, more challenging scenarios, where planning ahead is crucial. 
In the scenario in \Cref{7_cars} (from left to right), the RL agent (red), is in gap $g_0$, surrounded by slow vehicles.
In the starting position, there are 2 available gaps: stay in $g_0$ or start reaching $g_1$ 
\footnote{We do not consider double lane-changes directly, since there are more surrounding vehicles to be considered and it is harder to guarantee safety.}.
The RL agent and the leading vehicles of the reachable gaps are driving with the same speed, hence, there is no obvious short-term incentive to change a lane.
However, the options-selecting agent has learned that it is long-term beneficial to move over to the middle lane.
Since once it is there, the gap $g_2$ to the right opens up, where it will be able to speed up and overtake the slow vehicles.
For the high-level agent, since it is limited to \SI{3}{\second} lane-changes, the only available gap is the starting one.
This is because in the starting position its speed is the same as the speed of the leading vehicles and a \SI{3}{\second} lane change is not feasible.
In any given situation when for example the surrounding vehicles are driving with similar speed as the RL agent, and the RL agent is limited to discrete actions, a specific maneuver that may seems simple and logical, may be infeasible.
The IDM-based Sumo agent even though completely informed about the surroundings, without a mechanism for long-term considerations, failed to react appropriately. 
Since both reachable gaps $g_0$ and $g_1$ have the same speed, and moving to the next lane would result in an additional penalty, the greedy agent also fails to exit the initial gap.
In \Cref{3_cars}), the agent and the 3 surrounding vehicles drive with the same speed, lower than the RL agent's desired speed (\SI{30}{\meter/\second}).
However, the surrounding vehicles have achieved their desired velocities and have no intention to accelerate.
From that situation the agent has to break in order to reach the middle lane gap $g_1$.
Even though braking i.e. selecting a slow gap behind, translates directly into lower immediate reward, the options agent learned that it is essential in the long run, since staying in the initial gap is only locally optimal. \Cref{critical_scen} summarizes the performance of all agents averaged over the multiple runs, in terms of average speed and duration on the scenarios from \Cref{7_cars} (blue) and \Cref{3_cars} (green) respectively.

\section{CONCLUSION}\label{conclusion}

In this work, we introduced the Amortized Q-learning with Model-based Action Proposals approach for decision-making in common and critical highway situations. 
It is a Reinforcement Learning-based method, with an embedded optimal control trajectory planner, that it is not limited to fixed action sets and supports variable number of inputs. 
This is especially beneficial in confined space dynamic environments, where precise problem definition is crucial.
Moreover, it brings the best of both worlds together: a safe optimization-based trajectory planning and execution and long-term optimal decision-making.
We evaluated our options-selecting agent against 4 benchmark agents: a random agent, a greedy agent, IDM-based SUMO agent and a high-level discrete actions agent. 
We were able to show that the agents trained with our approach, not only achieved better overall performance but also were the only ones that managed to successfully navigate their way in the challenging scenarios.
In the future, it would be interesting to incorporate an additional interaction term in the reward signal such as courtesy.
This would make sure that the agent is cooperative and takes into account how its driving affects others. 

\printbibliography

@inproceedings{Montemerlo2009JuniorTS,
  title={Junior: The Stanford Entry in the Urban Challenge},
  author={M. Montemerlo and J. Becker and Suhrid Bhat and Hendrik Dahlkamp and D. Dolgov and S. Ettinger and D. H{\"a}hnel and T. Hilden and Gabe Hoffmann and Burkhard Huhnke and D. Johnston and S. Klumpp and D. Langer and A. Levandowski and Jesse Levinson and J. Marcil and D. Orenstein and Johannes Paefgen and I. Penny and A. Petrovskaya and Mike Pflueger and G. Stanek and David Stavens and A. Vogt and S. Thrun},
  booktitle={The DARPA Urban Challenge},
  year={2009}
}

@article{Urmson2008AutonomousDI,
  title={Autonomous driving in urban environments: Boss and the Urban Challenge},
  author={C. Urmson and J. Anhalt and J. Bagnell and Christopher R. Baker and R. Bittner and M. Clark and J. Dolan and David Duggins and Tugrul Galatali and C. Geyer and M. Gittleman and S. Harbaugh and M. Hebert and T. M. Howard and S. Kolski and A. Kelly and M. Likhachev and M. McNaughton and Nick Miller and K. M. Peterson and Brian Pilnick and R. Rajkumar and Paul E. Rybski and Bryan Salesky and Y. Seo and S. Singh and J. Snider and A. Stentz and W. Whittaker and Ziv Wolkowicki and Jason Ziglar and Hong Bae and T. Brown and Daniel Demitrish and B. Litkouhi and Jim Nickolaou and V. Sadekar and Wende Zhang and J. Struble and M. Taylor and M. Darms and D. Ferguson},
  journal={J. Field Robotics},
  year={2008},
  volume={25},
  pages={425-466}
}

@article{Bacha2008OdinTV,
  title={Odin: Team VictorTango's entry in the DARPA Urban Challenge},
  author={A. Bacha and C. Bauman and Ruel Faruque and Michael Fleming and Chris Terwelp and C. Reinholtz and Dennis W. Hong and A. Wicks and Thomas Alberi and D. Anderson and Stephen Cacciola and Patrick Currier and Aaron Dalton and Jesse Farmer and Jesse W. Hurdus and Shawn Kimmel and P. King and Andrew Taylor and David Van Covern and Mike Webster},
  journal={J. Field Robotics},
  year={2008},
  volume={25},
  pages={467-492}
}

@article{moritz_thrun,
  title={Optimal trajectory generation for dynamic street scenarios in a Fren{\'e}t Frame},
  author={M. Werling and Julius Ziegler and S. Kammel and S. Thrun},
  journal={2010 IEEE International Conference on Robotics and Automation},
  year={2010},
  pages={987-993}
}

@article{Schwarting2017ParallelAI,
  title={Parallel autonomy in automated vehicles: Safe motion generation with minimal intervention},
  author={W. Schwarting and Javier Alonso-Mora and Liam Pauli and S. Karaman and D. Rus},
  journal={2017 IEEE International Conference on Robotics and Automation (ICRA)},
  year={2017},
  pages={1928-1935}
}

@article{Borrelli2005MPCBasedAT,
  title={MPC-Based Approach to Active Steering for Autonomous Vehicle Systems},
  author={F. Borrelli and P. Falcone and T. Keviczky and J. Asgari and D. Hrovat},
  journal={International Journal of Vehicle Autonomous Systems},
  year={2005},
  volume={3},
  pages={265-291}
}

@article{Falcone2007PredictiveAS,
  title={Predictive Active Steering Control for Autonomous Vehicle Systems},
  author={P. Falcone and F. Borrelli and J. Asgari and H. E. Tseng and D. Hrovat},
  journal={IEEE Transactions on Control Systems Technology},
  year={2007},
  volume={15},
  pages={566-580}
}

@inproceedings{Schwarting2018PlanningAD,
  title={Planning and Decision-Making for Autonomous Vehicles},
  author={W. Schwarting and Javier Alonso-Mora and D. Rus},
  year={2018}
}

@article{Sutton1999BetweenMA,
  title={Between MDPs and Semi-MDPs: A Framework for Temporal Abstraction in Reinforcement Learning},
  author={R. Sutton and D. Precup and Satinder Singh},
  journal={Artif. Intell.},
  year={1999},
  volume={112},
  pages={181-211}
}

@misc{shalevshwartz2016safe,
      title={Safe, Multi-Agent, Reinforcement Learning for Autonomous Driving}, 
      author={Shai Shalev-Shwartz and Shaked Shammah and Amnon Shashua},
      year={2016},
      eprint={1610.03295},
      archivePrefix={arXiv},
      primaryClass={cs.AI}
}

@article{Treiber_2000,
   title={Congested traffic states in empirical observations and microscopic simulations},
   volume={62},
   ISSN={1095-3787},
   url={http://dx.doi.org/10.1103/PhysRevE.62.1805},
   DOI={10.1103/physreve.62.1805},
   number={2},
   journal={Physical Review E},
   publisher={American Physical Society (APS)},
   author={Treiber, Martin and Hennecke, Ansgar and Helbing, Dirk},
   year={2000},
   month={Aug},
   pages={1805–1824}
}

@misc{mnih2013playing,
      title={Playing Atari with Deep Reinforcement Learning}, 
      author={Volodymyr Mnih and Koray Kavukcuoglu and David Silver and Alex Graves and Ioannis Antonoglou and Daan Wierstra and Martin Riedmiller},
      year={2013},
      eprint={1312.5602},
      archivePrefix={arXiv},
      primaryClass={cs.LG}
}

@article{Silver2016MasteringTG,
  title={Mastering the game of Go with deep neural networks and tree search},
  author={D. Silver and Aja Huang and Chris J. Maddison and A. Guez and L. Sifre and George van den Driessche and Julian Schrittwieser and Ioannis Antonoglou and Vedavyas Panneershelvam and Marc Lanctot and S. Dieleman and Dominik Grewe and John Nham and Nal Kalchbrenner and Ilya Sutskever and T. Lillicrap and M. Leach and K. Kavukcuoglu and T. Graepel and Demis Hassabis},
  journal={Nature},
  year={2016},
  volume={529},
  pages={484-489}
}

@article{Silver2017MasteringTG,
  title={Mastering the game of Go without human knowledge},
  author={D. Silver and Julian Schrittwieser and K. Simonyan and Ioannis Antonoglou and Aja Huang and A. Guez and T. Hubert and L. Baker and Matthew Lai and A. Bolton and Yutian Chen and T. Lillicrap and F. Hui and L. Sifre and George van den Driessche and T. Graepel and Demis Hassabis},
  journal={Nature},
  year={2017},
  volume={550},
  pages={354-359}
}

@misc{wang2019deep,
      title={Deep Reinforcement Learning for Autonomous Driving}, 
      author={Sen Wang and Daoyuan Jia and Xinshuo Weng},
      year={2019},
      eprint={1811.11329},
      archivePrefix={arXiv},
      primaryClass={cs.CV}
}

@misc{wang2018reinforcement,
      title={A Reinforcement Learning Based Approach for Automated Lane Change Maneuvers}, 
      author={Pin Wang and Ching-Yao Chan and Arnaud de La Fortelle},
      year={2018},
      eprint={1804.07871},
      archivePrefix={arXiv},
      primaryClass={cs.RO}
}

@misc{wang2019quadratic,
      title={Quadratic Q-network for Learning Continuous Control for Autonomous Vehicles}, 
      author={Pin Wang and Hanhan Li and Ching-Yao Chan},
      year={2019},
      eprint={1912.00074},
      archivePrefix={arXiv},
      primaryClass={cs.LG}
}

@incollection{NIPS2017_6931,
title = {Deep Sets},
author = {Zaheer, Manzil and Kottur, Satwik and Ravanbakhsh, Siamak and Poczos, Barnabas and Salakhutdinov, Ruslan R and Smola, Alexander J},
booktitle = {Advances in Neural Information Processing Systems 30},
editor = {I. Guyon and U. V. Luxburg and S. Bengio and H. Wallach and R. Fergus and S. Vishwanathan and R. Garnett},
pages = {3391--3401},
year = {2017},
publisher = {Curran Associates, Inc.},
url = {http://papers.nips.cc/paper/6931-deep-sets.pdf}
}

@article{Mirchevska2018HighlevelDM,
	title={High-level Decision Making for Safe and Reasonable Autonomous Lane Changing using Reinforcement Learning},
	author={Branka Mirchevska and Christian Pek and Moritz Werling and Matthias Althoff and Joschka Boedecker},
	journal={2018 21st International Conference on Intelligent Transportation Systems (ITSC)},
	year={2018},
	pages={2156-2162}
}

@article{Hgle2019DynamicIF,
  title={Dynamic Input for Deep Reinforcement Learning in Autonomous Driving},
  author={Maria H{\"u}gle and G. Kalweit and Branka Mirchevska and M. Werling and J. Boedecker},
  journal={2019 IEEE/RSJ International Conference on Intelligent Robots and Systems (IROS)},
  year={2019},
  pages={7566-7573}
}

@article{Hgle2020DynamicIS,
  title={Dynamic Interaction-Aware Scene Understanding for Reinforcement Learning in Autonomous Driving},
  author={Maria H{\"u}gle and G. Kalweit and M. Werling and J. Boedecker},
  journal={2020 IEEE International Conference on Robotics and Automation (ICRA)},
  year={2020},
  pages={4329-4335}
}

@article{Kalweit2020InterpretableMT,
  title={Interpretable Multi Time-scale Constraints in Model-free Deep Reinforcement Learning for Autonomous Driving},
  author={G. Kalweit and Maria H{\"u}gle and M. Werling and J. Boedecker},
  journal={ArXiv},
  year={2020},
  volume={abs/2003.09398}
}

@misc{kalweit2020deep,
      title={Deep Inverse Q-learning with Constraints}, 
      author={Gabriel Kalweit and Maria Huegle and Moritz Werling and Joschka Boedecker},
      year={2020},
      eprint={2008.01712},
      archivePrefix={arXiv},
      primaryClass={cs.LG}
}

@book{Sutton1998,
  author = {Sutton, Richard S. and Barto, Andrew G.},
  edition = {Second},
  publisher = {The MIT Press},
  title = {Reinforcement Learning: An Introduction},
  url = {http://incompleteideas.net/book/the-book-2nd.html},
  year = {2018 }
}

@misc{vanhasselt2015deep,
      title={Deep Reinforcement Learning with Double Q-learning}, 
      author={Hado van Hasselt and Arthur Guez and David Silver},
      year={2015},
      eprint={1509.06461},
      archivePrefix={arXiv},
      primaryClass={cs.LG}
}

@article{sumo,
author = {Krajzewicz, Daniel and Erdmann, Jakob and Behrisch, Michael and Bieker-Walz, Laura},
year = {2012},
month = {12},
pages = {},
title = {Recent Development and Applications of SUMO - Simulation of Urban MObility},
volume = {3\&4},
journal = {International Journal On Advances in Systems and Measurements}
}

@INPROCEEDINGS{Watkins92q-learning,
    author = {Christopher J. C. H. Watkins and Peter Dayan},
    title = {Q-learning},
    booktitle = {Machine Learning},
    year = {1992},
    pages = {279--292}
}

@misc{alizadeh2019automated,
      title={Automated Lane Change Decision Making using Deep Reinforcement Learning in Dynamic and Uncertain Highway Environment}, 
      author={Ali Alizadeh and Majid Moghadam and Yunus Bicer and Nazim Kemal Ure and Ugur Yavas and Can Kurtulus},
      year={2019},
      eprint={1909.11538},
      archivePrefix={arXiv},
      primaryClass={cs.RO}
}

@INPROCEEDINGS{Hoel,
  author={C. {Hoel} and K. {Wolff} and L. {Laine}},
  booktitle={2018 21st International Conference on Intelligent Transportation Systems (ITSC)}, 
  title={Automated Speed and Lane Change Decision Making using Deep Reinforcement Learning}, 
  year={2018},
  volume={},
  number={},
  pages={2148-2155},
  doi={10.1109/ITSC.2018.8569568}}

@article{You,
author = {You, Changxi and Lu, Jianbo and Filev, Dimitar and Tsiotras, Panagiotis},
year = {2019},
month = {01},
pages = {},
title = {Advanced Planning for Autonomous Vehicles Using Reinforcement Learning and Deep Inverse Reinforcement Learning},
volume = {114},
journal = {Robotics and Autonomous Systems},
doi = {10.1016/j.robot.2019.01.003}
}

@article{enormousActions,
  author    = {Tom Van de Wiele and
               David Warde{-}Farley and
               Andriy Mnih and
               Volodymyr Mnih},
  title     = {Q-Learning in enormous action spaces via amortized approximate maximization},
  journal   = {CoRR},
  volume    = {abs/2001.08116},
  year      = {2020},
  url       = {https://arxiv.org/abs/2001.08116},
  archivePrefix = {arXiv},
  eprint    = {2001.08116},
  bibsource = {dblp computer science bibliography, https://dblp.org}
}

@article{DBLP:journals/corr/LiJDRT16,
  author    = {Lisha Li and
               Kevin G. Jamieson and
               Giulia DeSalvo and
               Afshin Rostamizadeh and
               Ameet Talwalkar},
  title     = {Efficient Hyperparameter Optimization and Infinitely Many Armed Bandits},
  journal   = {CoRR},
  volume    = {abs/1603.06560},
  year      = {2016},
  url       = {http://arxiv.org/abs/1603.06560},
  archivePrefix = {arXiv},
  eprint    = {1603.06560},
  bibsource = {dblp computer science bibliography, https://dblp.org}
}
\end{document}